\documentclass{article}

\usepackage{neurips_data_2024}

\usepackage{siunitx}
\usepackage{amssymb}
\usepackage[utf8]{inputenc} 
\usepackage[T1]{fontenc}    
\usepackage{url}            
\usepackage{booktabs}       
\usepackage{amsfonts}       
\usepackage{nicefrac}       
\usepackage{microtype}      
\usepackage{multirow} 
\usepackage{algorithm}
\usepackage{algorithmic}
\usepackage{graphicx}  
\usepackage{caption}   
\newcommand{\tzl}{\textcolor{black}}
\newcommand{\citep}{\cite}
\usepackage{booktabs}
\usepackage{amsmath}
\usepackage{amssymb}
\usepackage{mathtools}
\usepackage{amsthm}
\usepackage{MnSymbol}
\usepackage{wrapfig}
\usepackage{subfigure}
\usepackage{enumitem}
\setitemize{noitemsep,topsep=0pt,parsep=0pt,partopsep=0pt}
\usepackage{bm}
\usepackage{multicol}
\usepackage{multirow}

\theoremstyle{plain}

\theoremstyle{definition}

\theoremstyle{remark}

\usepackage{xr}
\usepackage{subfiles}

\usepackage{listings}
\usepackage{color}

\definecolor{dkgreen}{rgb}{0,0.6,0}
\definecolor{gray}{rgb}{0.5,0.5,0.5}
\definecolor{mauve}{rgb}{0.58,0,0.82}

\makeatletter
 \newcommand\figcaption{\def\@captype{figure}\caption}
 \newcommand\tabcaption{\def\@captype{table}\caption}
\makeatother

\title{NeoRL-2: Near Real-World Benchmarks for Offline Reinforcement Learning with Extended Realistic Scenarios}

\date{}

\author{%
    Songyi Gao$^{1}$\thanks{The first two authors contribute equally. $\diamond$ Correspondence to Yang Yu <yuy@nju.edu.cn>.}, Zuolin Tu$^{1}$,
	Rong-Jun Qin$^{1}$, Yi-Hao Sun$^{2}$, Xiong-Hui Chen$^{1,2}$, \textbf{Yang Yu}$^{1,2,\diamond}$\\
	$^{1}$\text{Polixir Technologies}\\
    $^{2}$\text{National Key Laboratory for Novel Software Technology, Nanjing University}\\
}

\begin{document}

\maketitle

\begin{abstract}
Offline reinforcement learning (RL) aims to learn from historical data without requiring (costly) access to the environment. To facilitate offline RL research, we previously introduced NeoRL, which highlighted that datasets from real-world tasks are often conservative and limited. With years of experience applying offline RL to various domains, we have identified additional real-world challenges. These include extremely conservative data distributions produced by deployed control systems, delayed action effects caused by high-latency transitions, external factors arising from the uncontrollable variance of transitions, and global safety constraints that are difficult to evaluate during the decision-making process. These challenges are underrepresented in previous benchmarks but frequently occur in real-world tasks. To address this, we constructed the extended Near Real-World Offline RL Benchmark (NeoRL-2), which consists of 7 datasets from 7 simulated tasks along with their corresponding evaluation simulators. Benchmarking results from state-of-the-art offline RL approaches demonstrate that current methods often struggle to outperform the data-collection behavior policy, highlighting the need for more effective methods. We hope NeoRL-2 will accelerate the development of reinforcement learning algorithms for real-world applications. The benchmark project page is available at \url{https://github.com/polixir/NeoRL2}.

\end{abstract}

\section{Introduction}

Reinforcement learning (RL) learns the optimal policy through trial-and-error in the environment, demonstrating excellent performance in various fields. From traditional game domains such as Go~\citep{silver2016alphago,silver2017alphazero} and Atari games~\citep{mnih2015human} to complex real-world tasks like recommender systems~\citep{2019VirtualTaobao} and sciences, such as Tokamak control~\citep{2022tokamak}, the application scope of RL continues to expand. However, when a fast and low-cost simulator is absent, conventional RL is hard to be applied~\citep{2021Gabriel}. This necessitates the need for offline RL~\citep{batchRL}. Offline RL learns a policy from static historical data, which alleviates the costly interaction with the environment.

Existing benchmarks for offline RL, such as D4RL~\citep{D4RL, 2021justin} and RL Unplugged~\citep{RLunplugged}, have significantly contributed to the development of offline RL. These benchmarks often focus on using historical data collected with different policies as offline datasets, and the tasks themselves are common benchmark tasks for online RL that may not fully capture the complexity of real-world application scenarios. For example, in real industrial systems, observations and rewards may contain significant delays due to sensor sampling and signal transmission \citep{1999Tan}. The data-collecting policies in real-world scenarios often utilize samples obtained from the operational control system without exploration. Real-world task scenarios are often not closed systems and are influenced by many external factors. Besides, the amount of historical data in the real world is small. The NeoRL~\citep{qinr2022neorl} benchmark has included conservative and limited datasets to reflect these two common properties in real-world datasets. Although NeoRL provides simulated domains from real-world tasks, such as industrial control and urban energy management, some critical challenges have not been contained. \tzl{In the real-world tasks we have experienced, there are effects of time delay, external factors, and safety constraints, which are also widely present in industry. For example, in a water treatment plant, the physical distance between pumps results in noticeable time delay effects for actions. There are delayed action effects when the environment has high latency and uncontrollable external factors that may affect the policy. When controlling drone flights, unpredictable winds in the environment can cause significantly different outcomes for the same control inputs, and these external factors are not influenced by the policy. Safety constraints are also prevalent in industrial environments; certain equipment must operate under specific conditions (e.g., temperature, speed), which restricts the action space. The operating policy should obey global safety constraints such that the offline datasets do not contain any unsafe behaviors. Thus, the offline algorithm has to learn a better and safer policy without knowing unsafe data.}

From these real-world scenarios, we have summarized some common characteristics: {\bf time delay}, {\bf external factors}, {\bf policy constraints}, {\bf data collected based on traditional control methods}, and {\bf limitations on data}. To further bridge the gap \tzl{and provide as many challenges as possible across different real-world scenarios, we extend NeoRL by proposing NeoRL-2. In NeoRL-2, we incorporated the above five common characteristics into seven tasks, making the benchmarks obtained from these tasks close to real-world scenarios.} It is important to note that we cannot directly use real business data in our benchmark due to infeasible real-world evaluation as well as data privacy. Thus, we purposely construct simulators for collecting datasets and evaluating policies. These simulators are not necessarily very sophisticated and realistic, but should exhibit important practical challenges for developing advanced algorithms.

NeoRL-2 includes new task scenarios that better reflect real-world task properties and includes traditional control methods as the data-collecting method. In summary, our contributions are as follows.

\begin{enumerate}[itemsep=0pt,topsep=0pt,parsep=0pt,leftmargin=*]
\item Tasks in NeoRL-2 cover a wider range of application domains, including robotics, aircraft, industrial pipelines, controllable nuclear fusion, healthcare, etc., encompassing key features such as delays, external factors, and safety constraints.

\item The data-collecting method in NeoRL-2 better aligns with real-world scenarios, employing deterministic methods for sampling. In some specific tasks, classical feedback controllers, such as the Proportional-Integral-Derivative (PID) controller~\cite{johnson2005pid}, are introduced.

\item We conducted experiments on these tasks using state-of-the-art (SOTA) offline RL algorithms and found that in most tasks, the trained policy of the current offline RL algorithms did not significantly outperform the behavior policy.
\end{enumerate}

By extending near real-world tasks in NeoRL-2, we hope that the development and implementation of RL in real-world scenarios can take into account these challenges and tackle more realistic domains.

\section{Related Work}

\subsection{Offline RL}

Offline RL focuses on learning from historical datasets. Unlike traditional online RL, offline RL does not require iterative interactions with the environment to update the policy. This requires algorithms to be capable of handling issues such as distributional shift \citep{2019Aviral} in the data, sparse rewards \citep{2022RengarajanVSKS}, and potential non-Markovian behavior \citep{lauri2023survey}.

Compared to online RL, a key advantage of offline RL is the ability to leverage existing historical data, which is highly valuable in many practical applications such as healthcare \citep{2006Damien,2017Niranjani,2018Komorowski}, industrial settings \citep{DENG2023221}, robotics \citep{2021Yevgen}, and autonomous driving \citep{2022Kiran}. In these domains, interacting with the environment is highly costly, risky, or impractical. Through offline RL, it is possible to learn and extract optimized policies from existing data without the need for additional online data collection.

Offline RL methods typically include model-free approaches and model-based approaches. Model-free methods directly learn policies or value functions from data without attempting to construct a dynamic model of the environment \citep{2018Sutton}. This includes techniques such as $Q$-learning and policy gradients. Model-based methods often start by learning a dynamics model of the environment from offline data and then use these models for planning or policy improvement through simulation \citep{luo2024survey}.

To address the challenges of offline RL, researchers have proposed various methods, primarily including policy constraint methods. These approaches minimize the impact of distributional shift by constraining the differences between the learning policy and the behavior policy \citep{workflow_offline_cql.2021, 2021td3bc}. For instance, they penalize actions that deviate from the original data distribution to prevent overestimation issues during the policy learning process \citep{yu2020mopo}. There are also methods based on uncertainty: These leverage the uncertainty inherent in the data to guide policy learning, balancing exploration and exploitation by estimating the uncertainty of the value function or the policy itself \citep{2023filter}. Regularization methods for the value function are another approach: These reduce the deviation of the learning policy from the original data distribution by incorporating regularization terms into the optimization process of the value function \citep{2021edac}. These methods aim to address key issues in offline RL, such as data quality and distributional shift issues.

\subsection{Offline RL Algorithm Benchmark}

Offline RL benchmarks play a crucial role in evaluating algorithms and advancements in the field by offering standardized environments and datasets \citep{levine2020offline}. Three notable benchmarks for testing and enhancing offline RL methods are D4RL \citep{D4RL}, RL Unplugged \citep{RLunplugged}, and NeoRL \citep{qinr2022neorl}, each with its unique focuses and dataset characteristics. 

D4RL is designed to mirror real-world challenges using datasets generated from various policies, including hand-designed controllers and human demonstrators. It covers properties like narrow and biased data distributions, sparse rewards, and sub-optimal data, facilitating existing offline algorithms and exposing potential flaws not evident in simpler tasks or datasets.
RL Unplugged provides a suite of datasets from domains such as games (e.g., Atari) and simulated motion control problems (e.g., DeepMind Control Suite). Emphasizing learning from offline datasets, it aims to overcome challenges like cost, safety, and ethical issues associated with online data collection. It provides clear evaluation protocols and reference performance baselines, facilitating researchers in comparing and analyzing different offline RL methods effectively. NeoRL focuses on the nature of datasets from real-world scenarios. It emphasizes conservative actions, limited data, and offline policy evaluation. By doing so, it aims to facilitate the development of offline RL algorithms better suited for real-world applications and promote the adoption of offline policy evaluation methods for improved policy selection before real-world deployment.

\section{The Reality Gap}
\label{sec.realitybap}
\subsection{Motivation of NeoRL-2}
\tzl{Previously, the NeoRL benchmark only considered the fundamental data properties from real-world tasks, where the data was collected by a conservative policy with small noise and the size of datasets was limited. In our practice of offline RL after releasing NeoRL, we encountered more difficulties that were not reflected in NeoRL or other offline RL benchmarks. The problems we face in these scenarios can be categorized as {\bf time delay}, {\bf external factors}, {\bf policy constraints}, {\bf data collected based on traditional control methods}, and {\bf limitations on data} (even more constrained than those in NeoRL).}

\tzl{
Since the real world is open and real-world tasks are very complex, it is impossible for the 7 task scenarios in the NeoRL-2 benchmark to cover all the challenges. We intend to summarize the challenges in real-world settings and to create simulators that more accurately reflect these challenges, rather than providing high-fidelity simulation environments. Thus, the most practical strategy is to reflect real-world issues as realistically as possible within the simulation environment. In addition, these properties indeed increase the difficulty of the task and lead to incomplete data. We hope that NeoRL-2 will encourage the community to pay more attention to these challenges when applying offline reinforcement learning to real-world tasks.
}

\subsection{Environment Properties}
The previously proposed benchmarks have played a significant role in advancing the development of offline RL techniques. However, based on our research, we have found notable gaps between real-world tasks and these benchmark tasks. Real-world tasks are generally more complex and challenging. Here, we articulate these properties that have been seen in real-world tasks but are frequently neglected by those existing benchmarks:

\begin{enumerate}[itemsep=0pt,topsep=0pt,parsep=0pt,leftmargin=*]

\item \textbf{Delay:} Delay can be caused by various factors, including delays in sensor sampling, signal transmission delays, and response delays. It is ubiquitous in different tasks and increases the uncertainty and learning difficulty of a task, which induces complex causality in understanding the system \citep{fridman2014introduction}. Delays can be observed not only in state transitions but also in obtaining rewards.

\item \textbf{External factors:} External factors are those that can affect the current system but are not influenced by the current internal system. Changes in these variables may modify the state distribution of the system environment, thus affecting the effectiveness of reward signals and policies \citep{2020Jinying}, further increasing the complexity and challenge of the task.

\item \textbf{Constrained control policies:} Real-world tasks often come with various constraints, which reveal physical limitations of the system, safety requirements, operational standards, or resource constraints \citep{2016Manuel}, where the datasets may only contain data that satisfies the constraints. Thus, due to the lack of exploration, offline algorithms can only learn from satisfactory data without explicitly knowing whether the current action will violate the constraints. 

\item \textbf{Data from traditional control methods:} In the real world, especially in the industrial sector, the control models in operation usually employ traditional control methods, such as PID. Data collected using these traditional methods tend to have a narrow distribution \citep{2022Tobias} and are difficult to model. In addition, these traditional methods often rely on feedback control, which may lead to learning incorrect transition relationships.

\item \textbf{Limitations on data availability:} In many real-world tasks, obtaining a large amount of training data may not be practical, leading to the problem of data insufficiency. In some special scenarios, only a few trajectories of data can be ideally collected, which poses a significant challenge to the optimization of offline policies \citep{levine2020offline, Mandlekar2021}. We further reduce the amount of datasets to match the real-world scenarios.

\end{enumerate}

Therefore, we propose NeoRL-2 as a supplement and improvement to existing benchmarks. The tasks in NeoRL-2 cover a wider range of real-world applications, including time-delay environmental transitions, external influencing factors, global constraints, data collected from traditional control methods, and limited data availability.

\tzl{Factors such as delays, external influences, constrained control policies, and data derived from traditional control methods significantly increase the task complexity and lead to data incompleteness. For instance, a large delay makes the action effect last an indefinitely long period, making the control policy non-Markovian. Policy constraints often restrict the data collection to only safe states and actions, resulting in incomplete space coverage. Additionally, external factors can introduce variability and may not be explicitly reflected in the dataset. All of these factors directly lead to the incompleteness of the observed information, gradually deviating from the standard MDP/POMDP assumptions. As many current offline reinforcement learning algorithms overlook these issues, these challenges could pose significant difficulties.}

\section{NeoRL-2 Tasks and Datasets}
In this section, we will provide a brief overview of all tasks included in NeoRL-2. Each task comprises a corresponding environment and dataset composition, with the dataset used for training policies and the environment used for policy testing.

\subsection{Tasks}

Next, we will provide a brief overview of the 7 tasks included in NeoRL-2. These tasks consist of a pipeline flow task and a human blood glucose concentration simulator, both of which exhibit significant time delay characteristics. Additionally, we have developed two additional environments: Rocket Recovery and Random Friction Hopper. In these environments, there are external factors that need to be included as part of the observations. To facilitate data collection using the PID control method, we constructed the Double Mass Spring Damper (DMSD) environment. Furthermore, the Safety Halfcheetah is a simulated task that imposes action constraints to ensure actions remain within a safe range. Finally, we have the Fusion environment, which represents the type of tasks in the real world where collecting offline data is extremely costly. In these tasks, the collection of offline data is particularly expensive, making the Fusion environment a representative example of such tasks.

{\bf Pipeline.} The Pipeline simulator models the flow of water through a 100-meter-long pipeline with a fixed velocity. The controller's objective is to regulate the flow rate by adjusting the inlet water gate. The simulation runs for 1000 time steps, with the target flow rate at the outlet influenced by an external policy and randomly selected from the set [50, 80, 110, 140]. At each time step, there is a 0.3\% probability of changing the target flow rate, averaging three changes per simulation.

{\bf Simglucose.}
The Simglucose simulator \citep{man2014uva} accurately simulates a diabetic patient's intricate blood glucose concentration variations throughout a day, starting at 8 a.m. The comprehensive simulator encompasses patient simulation, scenario simulation, sensor simulation, and insulin pump simulation. The patient randomly consumes meals and medication, with their effects gradually manifesting over time. The policy's primary objective is to safely control the medication regimen, ensuring that blood glucose levels remain within a desired range while accounting for the delayed effects of medication. This environment proves invaluable in optimizing and refining diabetes treatment strategies.

{\bf RocketRecovery.} The RocketRecovery simulator modifies the lunar\_lander\footnote{https://www.gymlibrary.dev/environments/box2d/lunar\_lander/} environment from Gym, incorporating wind force as an external factor. The task's objective is to land the rocket safely and as close to the target as possible, while maintaining safe angles and velocities. This environment presents a classic rocket trajectory optimization problem.

{\bf RandomFrictionHopper.} The RandomFrictionHopper simulator is a variation of the classic Hopper task, where a single-legged robot jumps and moves by controlling its leg actions. Each initialization sets a random ground friction coefficient in the range [1.5, 2.5], which is included as an extended dimension of the state observation. This variation introduces variability in the task's dynamics.

{\bf DMSD.} The Double Mass Spring Damper simulator involves two blocks with masses connected by springs and dampers. The task requires applying forces to control and stabilize the blocks at target positions. Notably, the interconnected springs and dampers mean that applying force to one block affects the other one. Each step in the simulation corresponds to 0.2 seconds in the real world, with a 100-step limit before automatic truncation.

{\bf SafetyHalfCheetah.} The SafetyHalfCheetah simulator is a variant of the HalfCheetah task, emphasizing safe high-speed running. In this task, the robot must sprint at maximum speed while ensuring its motion remains within safe limits to prevent accidents and damage. This task balances speed and safety, simulating real-world scenarios that require high-speed but safe robotic movement. Once the agent breaks the constraint, the environment will terminate and give a large negative score.

{\bf Fusion.} The Fusion simulator simulates the control of a Tokamak device for nuclear fusion. The objective is to control the Tokamak device solely based on observational states and attempt to stabilize the system at a target position during operation. However, the cost of conducting Tokamak device experiments in the real world is particularly high, making collected offline data in this environment valuable and data sparse. Due to the complexity of numerical simulations of Tokamak devices and the inherent error between numerical simulations and real device states, an LSTM neural network model has been utilized to "clone" the Tokamak device based on collected data \citep{Seo_2021}. We treat the "clone" model as a simulator of the Fusion task.

\begin{table}[ht]
\centering
\small
\caption{Comparison of the characteristics of each task.}
\label{table.taskcharacter}
\resizebox{1\textwidth}{!}{
\begin{tabular}{l|ccccc}
\toprule
& \multirow{3}{*}{Delay} & \multirow{3}{*}{External factors} & \multirow{3}{*}{\parbox{1.5cm}{\centering Constrained \\ control policies}} & \multirow{3}{*}{\parbox{2.5cm}{\centering Data from traditional \\ control methods}} & \multirow{3}{*}{\parbox{2.5cm} {Limitations on \\data availability}} \\
\\
\\

\midrule
Pipeline & $\checkmark$ & -- & -- & -- & -- \\
Simglucose & $\checkmark$  & $\checkmark$  & -- & -- & --  \\
RocketRecovery & -- &  $\checkmark$ & -- & -- & $\checkmark$ \\
RandomFrictionHopper & -- & $\checkmark$ & -- & -- & --\\
DMSD & -- & -- & -- & $\checkmark$ & -- \\
Fusion & -- & -- & -- &  -- & $\checkmark$ \\
SafetyHalfCheetah & -- & -- & $\checkmark$ & -- & $\checkmark$ \\
\bottomrule
\end{tabular}
}
\end{table}

Table \ref{table.taskcharacter} marks the characteristics possessed by each task with five attributes. These five attributes are detailed in Section \ref{sec.realitybap}.

\subsection{Datasets}

In real-world scenarios, data is often collected using conservative policies. For most environments, we use the Soft Actor-Critic (SAC) algorithm \citep{sac2018} for online policy training to obtain high-performance policies. During training, we record the policy model after each round and rank them based on their performance. Considering that real-world methods are often non-optimal policies, we randomly select policies ranked within the 20-75\% range as sampling policies. In the DMSD environment, we employ the Bayesian Optimization parameter search method \citep{Fernando2014} to fine-tune the three key parameters of the PID controller. We rank all PIDs based on their performance and select the PID with a performance ranking of 56\% as the offline data sampling policy for the DMSD dataset.

After identifying the sampling policies, we collect data in all built-in environment simulators. \tzl{In most real-world tasks, there is typically only one policy for collecting offline datasets, leading to a lack of diversity. So, in NeoRL-2 it is important to clarify that all policies are deterministic when interacting with the environment to collect data.} The collected data are then divided into training and validation datasets, with 100,000 training dataset samples and 20,000 for validation for each environment. Notably, the Fusion environment has a unique data sampling process due to the high cost of controlled nuclear fusion experiments. To reflect this, we limit the number of experimental trajectories, resulting in a training dataset of 2,000 samples and a validation dataset of 500 samples. Similarly, limited data availability is observed in the Rocket Recovery and Safety Halfcheetah environments, as detailed in Table \ref{task_configuration} of Appendix \ref{sec.simulator}.

\section{Experiments}

\subsection{Comparing Methods}
We validated the following methods in the tasks of NeoRL-2.

\subsubsection{Baseline}
{\bf Expert Policy.} Specifically, we use the soft actor-critic (SAC)~\citep{sac2018} algorithm to perform online policy improvement for each environment in NeoRL-2. The model achieving the highest score will be saved. It is important to note that the highest-scoring policy obtained by SAC represents a good policy, but not necessarily the optimal one. \textbf{In normalization of data scores (0-100), the scores given by expert policies are used as the maximum bound.}

{\bf Random Policy.} We employ uniform random sampling to generate actions within the action space of each environment. The random policy is generally considered a poor policy, often demonstrating terrible performance. \textbf{The performance of this policy in the environment is typically used as the minimum bound for the normalization of data scores (0-100).}

{\bf Behavior Policy.} This approach involves preserving sub-optimal policies generated during the SAC-training of expert policies. In the real world, most offline data collection does not achieve optimal policy performance, resulting in offline datasets. If some environments already had optimal policies, there would be no need to apply RL for performance improvement. Thus, we compile datasets from the policies saved during SAC training and record the average return value of the trajectories in these datasets. Note that the dataset for DMSD is collected using a PID policy.

\subsubsection{Model-Free Methods}
To establish a comprehensive benchmark, we include several representative algorithms for offline policy improvement. These include the Behavior Cloning (BC) method, which uses supervised learning to replicate the policies from the data. We also implement offline RL methods such as CQL \citep{2020cql}, EDAC \citep{2021edac}, MCQ \citep{2022mcq}, and TD3BC \citep{2021td3bc}. All methods implementations can be viewed at \url{https://github.com/polixir/OfflineRL}.

\begin{itemize}[itemsep=0pt,topsep=0pt,parsep=0pt,leftmargin=*]
    \item \emph{CQL}~\citep{2020cql}.
    Conservative $Q$-Learning (CQL) method prevents overestimation of the state-action value function $Q$ by especially penalizing out-of-distribution (OOD) data points. By obtaining a conservative $Q$ function, CQL minimizes the difference between the current $Q$-values and the true values in the data, while training the policy by maximizing the $Q$ function. This approach, applied within the SAC framework, effectively mitigates the issue of $Q$-value overestimation, resulting in a more reliable offline policy improvement algorithm.
    
    \item \emph{EDAC}~\citep{2021edac}. Ensemble-Diversifying Actor-Critic (EDAC) method builds upon the SAC algorithm by utilizing the clipped $Q$-learning method, commonly used in online RL, to effectively penalize OOD data points. The clipped $Q$-learning method computes a pessimistic estimate by choosing the minimum of $Q$-values. Additionally, EDAC significantly enhances the algorithm's effectiveness by increasing the number of $Q$-networks, which is also proved beneficial in various actor-critic algorithms. 
    By incorporating Ensemble Gradient Diversification, EDAC effectively reduces the number of $Q$-networks while maintaining, and even surpassing, the effectiveness of state-of-the-art (SOTA) algorithms.

    \item \emph{MCQ}~\citep{2022mcq}. Mildly Conservative $Q$-learning (MCQ) method aims to reduce the overestimation of the $Q$ function, particularly by maintaining conservatism for OOD actions. When adding a penalty term for OOD data, the value of this penalty is based on assigning appropriate scores to OOD data within the constructed $Q$-network, rather than simply increasing the penalty strength. Experimental results show that this conservative approach does not significantly cause overestimation in the $Q$ network and achieves commendable performance.
    
    \item \emph{TD3BC}~\citep{2021td3bc}. TD3BC makes slight modifications to existing RL algorithms but achieves significant performance improvements. It modifies the policy gradient component of the Twin Delayed Deep Deterministic Policy Gradient (TD3) algorithm \citep{2018Fujimoto}. For the objective function, the state-action value function result subtracts the Mean Squared Error (MSE) between the current policy network output and the action from the dataset. Essentially, TD3BC adds a BC regularization term to the standard policy update of TD3 to encourage the policy to favor actions present in the training dataset.
    
\end{itemize}

\subsubsection{ Model-Based Methods}
The advantages of model-based offline RL over model-free methods include higher data utilization efficiency, faster convergence speed, and great explainability. By utilizing offline data for environment model training, model-based methods can effectively make use of the available data. We employ model-based RL methods, including MOPO \citep{yu2020mopo}, COMBO \citep{2021combo}, RAMBO \citep{2022rambo}, and Mobile \citep{2023mobile}. These methods construct a dynamic model from the data and use this model to enhance policies.

\begin{itemize}[itemsep=0pt,topsep=0pt,parsep=0pt,leftmargin=*]
    \item \emph{MOPO}~\citep{yu2020mopo}. Model-based Offline Policy Optimization focuses on the distribution shift problem between the trained dynamic model and the real environment in model-based RL. To mitigate this issue, during subsequent policy learning, the rewards obtained from interactions between the policy network and the model are augmented with a penalty that accounts for the discrepancy between the model and the real data. This ensures that the policy, when making action choices, considers the extent of deviation from real environment transitions. By aligning the policy as closely as possible with the transition dynamics in the data, MOPO aims to optimize the policy effectively.
    \item \emph{COMBO}~\citep{2021combo}. 
    Conservative Offline Model-Based Policy Optimization extends MOPO by additionally addressing potential inaccuracies in calculating the discrepancy between the model and the real data, as well as poor performance in some scenarios. During the training of the value network, COMBO incorporates both offline datasets and model-generated data, introducing an additional regularization term to generated out-of-support state-action pairs. This approach enables a conservative estimate of the value function without the need to directly measure the error between generated data and offline data, thereby enhancing the robustness and performance of the algorithm.
    \item \emph{RAMBO}~\citep{2022rambo}. Robust Adversarial Model-Based Offline RL constructs the dynamic model using the BC method. Based on that, RAMBO performs model-based policy improvement, primarily using the SAC algorithm structure. Notably, RAMBO samples from a buffer that combines training data with trajectory data generated by the model. During each update cycle, the model also undergoes updates. By utilizing maximum likelihood estimates to adjust the model, RAMBO influences the loss function. This approach ensures that the model continuously approximates the environment transitions present in the training data throughout the policy training process.
    \item \emph{Mobile}~\citep{2023mobile}. To ensure conservatism in its usage, Mobile incorporates Model Bellman Inconsistency as a measure of uncertainty to assess the discrepancy between generated data and offline data. This metric allows for quantitative analysis of the inconsistency of Bellman estimations produced by the ensemble of dynamic models. In regions where there is abundant offline data, Bellman estimations typically exhibit small errors and low discrepancies between estimations. However, in regions with rare data, there is a higher inconsistency between Bellman estimations. Mobile addresses actions with poor consistency of Bellman estimations by applying larger penalties, further ensuring that the policy avoids risky actions in such regions.
\end{itemize}

\subsection{Benchmarking Results}
\label{sec.bm_result}

\begin{figure}[htbp] 
\centering 
\includegraphics[width=1\textwidth]{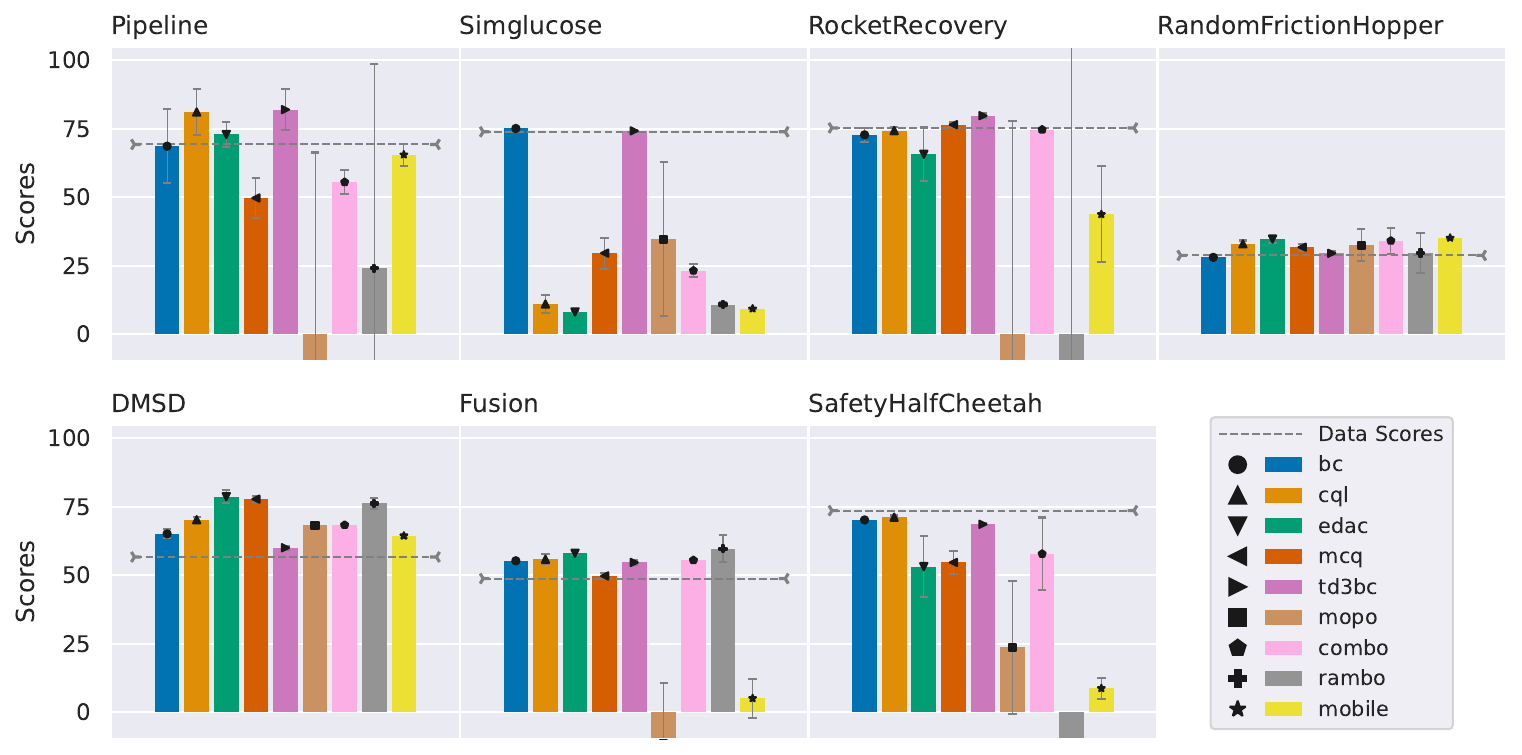} 
\caption{The baseline algorithm normalized scores over seven environments compared to that of offline data.}
\label{fig.resultshist} 
\end{figure}
As there are no very stable and consistent offline policy evaluation methods~\cite{OffPolicyEval,qinr2022neorl}, we only conduct online policy evaluation. We run each configuration of the hyperparameters with 3 random seeds and choose the policy at the final training stage to conduct the online test. We report the results from the best hyperparameter across 3 seeds and the 3 seeds are the same for all the algorithms.
Figure \ref{fig.resultshist} illustrates the normalized scores obtained by the baseline algorithm, with each color and symbol representing a different baseline algorithm as indicated in the legend. This visualization, derived from Table \ref{table.algo_final}, offers an intuitive overview of the performance of tuned hyper-parameters that yield the highest mean score among the current algorithms. The performance of each algorithm on specific tasks is computed using three random seeds, and their standard errors are depicted with error bars. From an environmental perspective, the performance of almost all baseline algorithms surpasses the scores in the offline data in the RandomFrictionHopper and DMSD environments. Model-based algorithms demonstrate lower stability compared to model-free algorithms, particularly evident in the RocketRecovery, Fusion, and SafetyHalfCheetah tasks, where MOPO and RAMBO exhibit larger error bars. This highlights the heightened sensitivity of MOPO and RAMBO algorithms to random seed selection in these tasks, while other algorithms showcase smaller error bars, indicating better stability. In terms of scores, no algorithm surpasses 95 in any task \footnote{A normalized score of 95 can be treated as having solved the task.}. The TD3BC algorithm achieves the highest score of 81.95 points in the Pipeline task but falls short of 95 points. Consequently, we have yet to identify any algorithm capable of successfully solving the tasks in NeoRL-2, and there has been no significant improvement in scores compared to those in the data.

\begin{table}[ht]
\centering
\small
\setlength{\tabcolsep}{6pt} 
\renewcommand{\arraystretch}{1.5} 
\caption{The parameter configuration that achieves the highest score in the baseline algorithm over all 7 environments. $\pm$ denotes standard error over three random seeds.}
\label{table.algo_final}
\resizebox{1\textwidth}{!}{
\begin{tabular}{l|c r@{~$\pm$~}l r@{~$\pm$~}l r@{~$\pm$~}l r@{~$\pm$~}l r@{~$\pm$~}l r@{~$\pm$~}l r@{~$\pm$~}l r@{~$\pm$~}l r@{~$\pm$~}l}
\toprule
& \textbf{Data} & \multicolumn{2}{c}{\textbf{BC}} & \multicolumn{2}{c}{\textbf{CQL}} & \multicolumn{2}{c}{\textbf{EDAC}} & \multicolumn{2}{c}{\textbf{MCQ}} & \multicolumn{2}{c}{\textbf{TD3BC}} & \multicolumn{2}{c}{\textbf{MOPO}} & \multicolumn{2}{c}{\textbf{COMBO}} & \multicolumn{2}{c}{\textbf{RAMBO}} & \multicolumn{2}{c}{\textbf{MOBILE}} \\
\midrule
\textbf{Pipeline} & 69.25 & 68.61 & 13.42 & 81.08 & 8.25 & 72.88 & 4.64 & 49.7 & 7.39 & \bf{81.95} & \bf{7.46} & -26.33 & 92.65 & 55.50 & 4.28 & 24.06 & 74.39 & 65.51 & 4.05 \\
\textbf{Simglucose} & 73.87 & \bf{75.07} & \bf{0.71} & 10.99 & 3.35 & 8.11 & 0.31 & 29.57 & 5.65 & 74.21 & 0.38 & 34.64 & 28.13 & 23.18 & 2.47 & 10.8 & 0.86 & 9.29 & 0.17 \\
\textbf{RocketRecovery} & 75.27 & 72.75 & 2.53 & 74.32 & 1.44 & 65.65 & 9.81 & 76.46 & 0.81 & \bf{79.74} & \bf{0.91} & -27.66 & 105.61 & 74.66 & 0.70 & -44.21 & 263.00 & 43.71 & 17.53 \\
\textbf{RandomFrictionHopper} & 28.73 & 28.03 & 0.35 & 32.96 & 1.23 & 34.69 & 1.29 & 31.74 & 1.25 & 29.51 & 0.66 & 32.48 & 5.79 & 34.08 & 4.74 & 29.62 & 7.15 & \bf{35.12} & \bf{0.47} \\
\textbf{DMSD} & 56.64 & 65.09 & 1.64 & 70.24 & 1.12 & \bf{78.65} & \bf{2.30} & 77.75 & 1.16 & 60.01 & 0.79 & 68.23 & 0.65 & 68.33 & 0.41 & 76.22 & 1.92 & 64.36 & 0.78 \\
\textbf{Fusion} & 48.75 & 55.24 & 0.30 & 55.86 & 1.89 & 58.00 & 0.72 & 49.70 & 1.05 & 54.64 & 0.82 & -11.60 & 22.20 & 55.48 & 0.28 & \bf{59.63} & \bf{4.99} & 5.02 & 7.05 \\
\textbf{SafetyHalfCheetah} & \bf{73.56} & 70.16 & 0.38 & 71.18 & 0.61 & 53.11 & 11.14 & 54.65 & 4.32 & 68.58 & 0.39 & 23.71 & 24.28 & 57.79 & 13.27 & -422.43 & 307.49 & 8.68 & 3.93 \\
\bottomrule
\end{tabular}
}
\end{table}
Table \ref{table.algo_final} presents the highest scores of all baseline algorithms with tuned hyper-parameters in the NeoRL-2 tasks, along with the standard errors computed from three random seeds. From Table~\ref{table.algo_final}, it is evident that in the SafetyHalfCheetah task, none of the algorithms exceed the scores in the offline data, with RAMBO exhibiting particularly high standard errors, indicating significant instability. The most notable improvement in scores is observed in the DMSD task, where the EDAC algorithm's performance increases to 78.56, representing an improvement of about 22 scores.

\begin{figure}[h]
    \centering
    \begin{minipage}[b]{0.45\textwidth}
        \centering
        \includegraphics[width=\textwidth]{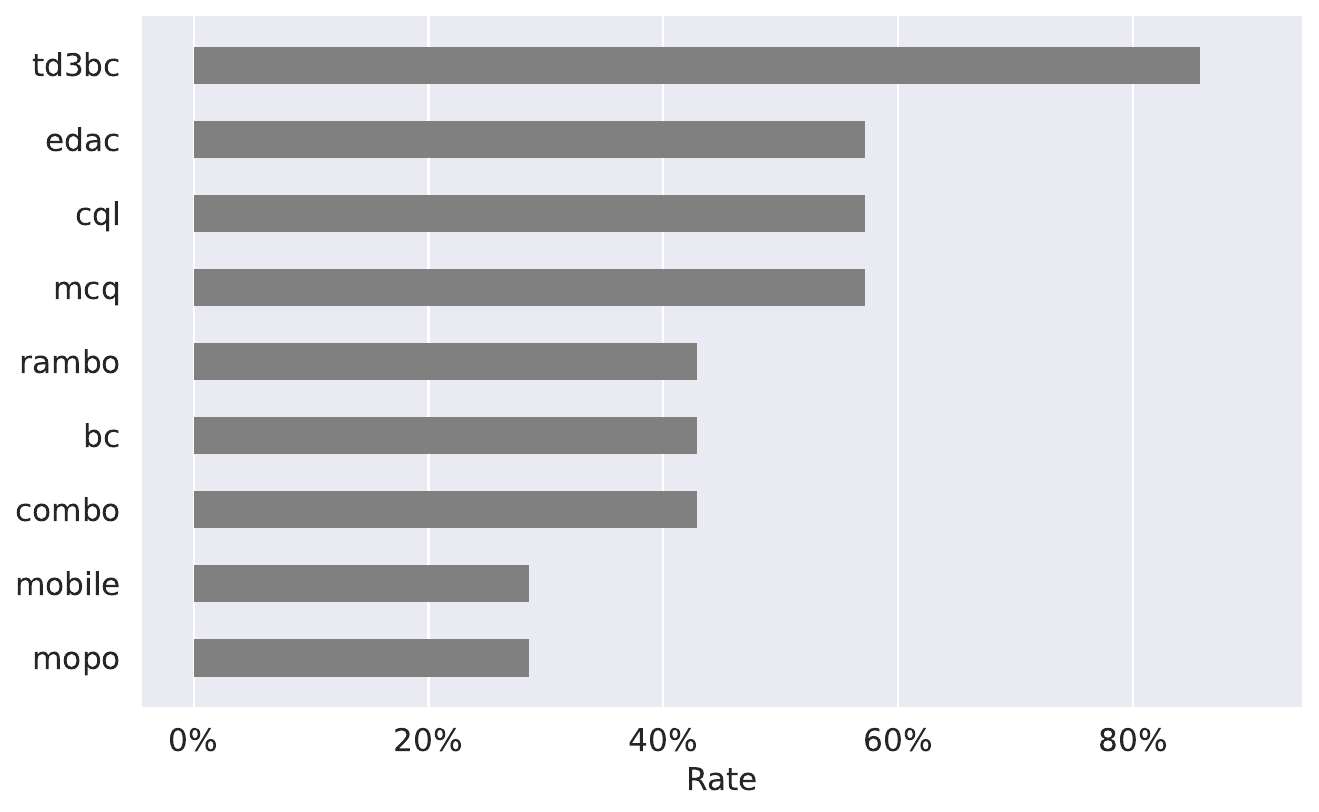} 
        \caption{The proportion of baseline algorithms whose scores exceed the data scores.}
        \label{fig.resultrate}
    \end{minipage}
    \hfill
    \begin{minipage}[b]{0.45\textwidth}
        \centering
        \begin{tabular}{l|cccc}
        \toprule
         & +0 & +3 & +5 & +10 \\
        \midrule
        BC & 3 & 2 & 2 & 0 \\
        CQL & 4 & 4 & 3 & 2 \\
        EDAC & 4 & 4 & 3 & 1 \\
        MCQ & 4 & 2 & 1 & 1 \\
        TD3BC & 6 & 4 & 2 & 1 \\
        MOPO & 2 & 2 & 1 & 1 \\
        COMBO & 3 & 3 & 3 & 1 \\
        RAMBO & 3 & 2 & 2 & 2 \\
        MOBILE & 2 & 2 & 2 & 0 \\
        \bottomrule
        \end{tabular}
        \tabcaption{The specific number of tasks in which an algorithm surpasses data scores.}
        \label{table.scoresover}
    \end{minipage}
\end{figure}

Figure \ref{fig.resultrate} compares the proportion of baseline algorithms that exceed data scores in seven tasks. The results show that the TD3BC algorithm surpasses data scores in more than 80\% of the tasks. Additionally, the proportion of successes is higher for model-free algorithms compared to model-based ones. It is worth mentioning that the proportion calculated here is based on the results from Table \ref{table.algo_final}, where any algorithm's score surpassing the data score is counted as a successful improvement. This approach may result in cases where the scores are close but the improvement is not significant.

Table \ref{table.scoresover} specifically lists the number of times each baseline algorithm exceeds the scores in the data. Here, +0, +3, +5, and +10 represent the thresholds for exceeding the scores in the data. For example, +10 means that the algorithm's score exceeds the data score by more than 10 points. With these results, we can gain a more intuitive understanding of the degree of improvement for each algorithm. From the results, RAMBO and CQL achieved a 10-score improvement in two out of seven tasks. However, none of the algorithms can achieve a 5-score improvement in more than half of the tasks. These findings suggest that current baseline algorithms have not significantly improved performance in the NeoRL-2 tasks. While there may be improvements in some tasks, we have not observed any algorithm that can achieve a 5-point improvement in most tasks. This indirectly indicates that in the tasks where these SOTA algorithms achieve significant performance improvements, the tasks are relatively simple. Once these tasks are replaced by the NeoRL-2 tasks, which are closer to the real world, the performance of these algorithms is far from the performance they claim.

\section{Conclusion}
Offline RL aims to learn from historical data, eliminating the need for new data collection in real-world environments. To evaluate the effectiveness of offline RL algorithms, various benchmark suites have been proposed. However, existing benchmarks often fail to fully capture the critical characteristics of real-world tasks. To address this, we introduce NeoRL-2, a benchmark suite that encompasses tasks in robotics, aerial vehicles, industrial pipelines, nuclear fusion, and the medical fields. These tasks incorporate features such as delays, external factors, and constraints, making NeoRL-2 more representative of real-world scenarios. We employ a deterministic method for sampling and restrict the number of samples, aligning with realistic task settings. Additionally, in certain specialized scenarios, we incorporate the classic PID controller as a data sampling method. We tested SOTA offline RL algorithms using NeoRL-2 and analyzed the results. Our experimental results indicate that, in most tasks, these algorithms do not significantly outperform the policies used for data collection. Through this work, we aim to promote a closer integration of offline RL algorithm research with real-world applications, facilitating the practical implementation of RL in real-world scenarios.

\tzl{NeoRL-2 is designed to provide a range of easy-to-use simulation environments for evaluating offline reinforcement learning algorithms, however, there are still gaps between the benchmarking results and the real-world performances of current offline RL algorithms. The simulation environments still cannot fully replicate the complexity of real-world tasks, especially when the transition of some real-world tasks is unclear. This is also a common issue faced by the offline RL community, and we believe NeoRL-2 will facilitate the development of both future benchmark tasks and offline algorithms.}

The revised LaTeX content includes grammar corrections and improvements in sentence structure for better clarity and readability. The technical content remains intact. Please let me know if you have any further questions or require additional assistance.

\begin{ack}
This work was supported by the National Science and Technology Major Project (Grant No. 2022ZD0114805). The authors also thank Jiawei Chen for participating in the early version of the datasets. 
\end{ack}

\bibliography{NeoRL_benchmark}
\bibliographystyle{unsrt}

\newpage
\appendix

\section{Additional Discussion}\label{Appendix.add_discuss}

\subsection{Experiment Discussion}
\tzl{According to the results shown in Section \ref{sec.bm_result}, factors such as delay, external factors, constrained control policies, and data from traditional control methods indeed increase task difficulty and lead to incomplete data. These factors lead to data limitations and the emergence of non-MDP properties in the task itself, which might be the reasons why the aforementioned SOTA algorithms fail to achieve better results.}

\tzl{In the future, improvements to offline RL algorithms could potentially be made in the following areas. For instance, addressing time delays could involve using RNN architecture or designing algorithms that account for latency. Tasks influenced by external factors can be analyzed using causal analysis techniques. Policy constraints may be managed through constrained MDPs or safe RL methods \citep{2022Shangding}. It is important to note that in offline environments, solutions like safe RL may not be directly applicable, as real-world data might lack unsafe scenarios, unlike the exploration of unsafe behavior in standard safe RL training. Previously, \citep{2022COptiDICE} presented an offline constrained reinforcement learning algorithm that optimizes policies in stationary distributions, improving constraint-based tasks. Recently, \citep{quan2024learning} has used inverse reinforcement learning to extract feasible constraints from superior distributions, highlighting policies that surpass expert reward maximization. It might be worthwhile to use the methods from these two works to address the tasks involving constrained control policies in NeoRL-2, while the limited safe data could present new challenges. Due to current challenges, we have yet to identify practical methods for implementation.
}

\section{Task Simulators and Datasets}
\label{sec.simulator}

\begin{table}[!h]
	\centering
        \renewcommand{\arraystretch}{1.3}
	\caption{Configuration of task simulators and datasets.}
	\label{task_configuration}
	\scalebox{0.9}{\begin{tabular}{cccccccccc}
			\toprule
			Environment & \begin{tabular}[c]{@{}c@{}}Observation \\ Shape\end{tabular} & \begin{tabular}[c]{@{}c@{}}Action \\ Shape\end{tabular} & \begin{tabular}[c]{@{}c@{}}Have \\ Done\end{tabular} & \begin{tabular}[c]{@{}c@{}}Max \\ Timesteps\end{tabular} & \begin{tabular}[c]{@{}c@{}}Training Dataset \\ Samples\end{tabular} & \begin{tabular}[c]{@{}c@{}}Validation Dataset \\ Samples\end{tabular} \\
			\midrule
                Pipeline & 52 & 1 & False & 1000 & 100000 & 20000 \\ 
                Simglucose & 31 & 1 & True & 480 & 100257 & 20018 \\ 
                RocketRecovery & 7 & 2 & True & 500 & 19977 & 3491 \\ 
                RandomFrictionHopper & 13 & 3 & True & 1000 & 100105 & 20047 \\ 
                DMSD & 6 & 2 & False & 100 & 100000 & 20000 \\ 
                Fusion & 15 & 6 & False & 100 & 2000 & 500 \\ 
                SafetyHalfCheetah & 18 & 6 & False & 1000 & 18000 & 2000 \\ 
			\bottomrule
	\end{tabular}}
\end{table}

{\bf Pipeline}. In the pipeline flow control simulation environment, the observation space is a 52-dimensional vector comprising the actual and target water flows at the pipeline's end, along with the water flow and action values observed over the past 25 time steps. The action space is a continuous 1-dimensional variable, allowing adjustments to the water flow at the pipeline's start within a range of [-1, 1]. Due to the pipeline's length, there is a delay between actions and their impact on rewards, necessitating the consideration of temporal dynamics. Consequently, an effective control policy must account for these reward delays. The simulation is capped at 1000 time steps; exceeding this limit results in automatic termination of the trajectory. 

{\bf Simglucose}.
The Simglucose simulator is an environment that simulates the variations in the blood glucose concentration of a diabetic patient throughout one day. The state space consists of 31 dimensions, including the current subcutaneous blood glucose concentration measured by sensors (1 dimension by default) and the physiological attributes of each patient (30 dimensions). The action space is 1-dimensional and represents the basal insulin dose to be delivered to the insulin pump. The reward function is based on the difference between the actual subcutaneous blood glucose concentration at the current and previous time steps, taking into account the cases of hypoglycemia (<40) or hyperglycemia (>=600) to encourage maintaining blood glucose within a safe range. There are two termination conditions: when the current subcutaneous blood glucose concentration falls below 40 or exceeds 600, and when the trajectory length exceeds 480 steps, which corresponds to a simulation of 24 hours. The delay in the task is influenced by many factors and is a dynamic value. The environment is affected by 30 external variables related to patient attributes. The source code for \texttt{lunar\_lander} can be found on GitHub at: \url{https://github.com/jxx123/simglucose}.

{\bf RocketRecovery}.
The RocketRecovery is a simulator based on the \texttt{lunar\_lander} environment in Gym, with the addition of external factors called wind. The objective is to land the rocket safely at the target position with a proper angle and velocity, making it a classic rocket trajectory optimization problem. The state space includes the horizontal position, vertical position, horizontal velocity, vertical velocity, orientation angle, angular velocity, and wind. The action space has two dimensions, which are used to control the thrust of the main engine and the left and right engines. The reward function is calculated based on the distance between the rocket and the landing platform, the number of times the thrusters are fired (both main and side engines), and whether the landing is safe. It includes distance rewards, engine firing penalties, and landing rewards. The termination conditions include exceeding 500 timesteps, indicating a timeout for the trajectory, or the rocket's Y coordinate being less than 0.05, indicating a successful landing. The initial state of the rocket is at the top center with random initial forces and velocities. The number of data points is limited to only 100, with the longest trajectory having 500 steps and the shortest having 77 steps, averaging 200 steps. The data is influenced by external variables, specifically wind speed, which is not fixed; the distribution of wind speed in the data ranges from [-9.64, 9.64]. The source code for \texttt{lunar\_lander} can be found on GitHub at: \url{https://github.com/openai/gym/blob/master/gym/envs/box2d/lunar_lander.py}.

{\bf RandomFrictionHopper}. 
The RandomFrictionHopper simulator is a variant of the Mujoco Hopper, which is a robot with a single leg used for jumping and movement. In the RandomFrictionHopper task, the ground friction coefficient is randomly set within the range of [1.5, 2.5] each time the environment is initialized. The friction coefficient is an external factor that can be observed in the state. The state space is 13-dimensional and describes the current state of the robot in the RandomFrictionHopper task. It includes information such as the robot's position, velocity, joint angles, joint angular velocities, and ground friction coefficient. These variables provide details about the robot's position and motion, enabling it to make decisions. The action space is 3-dimensional, allowing the robot to control its joints by applying different forces or torques. This affects the robot's velocity and posture, influencing its movements. The reward function consists of two components: a velocity reward that encourages fast running by providing a positive reward based on the robot's forward velocity, and an energy cost penalty that penalizes large joint torques or forces. This encourages the robot to use smaller forces or torques for its movements. The termination conditions include two aspects: if the trajectory exceeds 1000 time steps, it ends due to the time limit, or if the robot enters an unsafe state where certain state indicators such as the z position or angle exceed the predefined safety range. The maximum number of steps is 1000, and exceeding this limit results in the trajectory ending due to the time constraint. The source code for Mujoco Hopper can be found on GitHub at: \url{https://github.com/openai/gym/blob/master/gym/envs/mujoco/hopper.py}.

{\bf DMSD}.
The state space of the DMSD (Double Mass Spring Damper) simulator consists of two parts: one part includes the individual states of the two sliders, including their positions and velocities, and the other part reflects the x-axis coordinates of the target positions for the two sliders. The state space has a total of 6 dimensions. The action space is a 2-dimensional array that represents the magnitude of the forces applied to the sliders. Negative values indicate forces applied to the left, while positive values indicate forces applied to the right. The system normalizes the forces, so the action space ranges from -1 to 1. The data was sampled using a PID control strategy. There are 200 trajectories in the dataset, each with a length of 100 steps. The source code for DMSD can be found on GitHub at: \url{https://github.com/IanChar/GPIDE/blob/main/rlkit/envs/double_msd_env.py}.

{\bf Fusion}. The Fusion simulator is a Tokamak device simulation system used for controlling controllable nuclear fusion devices. The state space is 15-dimensional and includes physical quantities such as plasma current, plasma elongation, the shape difference of the upper and lower halves of the plasma elliptical cross-section, the inner and outer radii of the plasma separation surface in the midplane, as well as the power of three ion injectors, and values related to poloidal plasma beta, interface flow at 95\% position, and internal inductance coefficient. The action space is a 6-dimensional continuous space with values ranging from 0 to 1. The reward function is calculated based on the difference between the target state and the new state when the agent takes an action. The maximum number of steps in this environment is 100, representing 200 seconds in the physical world, after which the environment is automatically truncated. The dataset only has 20 trajectories, with each entry having 100 steps. The source code for Fusion can be found on GitHub at: \url{https://github.com/jaem-seo/AI_tokamak_control}.

{\bf Safety HalfCheetah}. The SafetyHalfCheetah simulator is a variant of the MUJOCO HalfCheetah designed to make the robot run at maximum speed while maintaining safety. The state space is 18-dimensional and describes the current state of the robot, including position, velocity, joint angles, and joint angular velocities. The action space is 6-dimensional and defines the actions that the robot can take, typically continuous values such as joint forces or torques. The reward function consists of three components: velocity reward, which provides a positive reward based on the forward velocity of the robot to encourage fast running; energy cost penalty, which penalizes the magnitude of joint torques or forces to encourage movements with smaller forces or torques; and a penalty for exceeding the safety speed boundary (3.2096), which results in a penalty of -100. The maximum number of steps is 1000 time steps, and the trajectory automatically terminates if this limit is exceeded. There are 18 trajectories in the dataset, each with a length of 1000 steps. The source code for MUJOCO HalfCheetah can be found on GitHub at: \url{https://github.com/openai/gym/blob/master/gym/envs/mujoco/half_cheetah.py}.

Each task's data is divided into training and validation datasets, with the validation dataset typically comprising about one-fourth of the training dataset's samples. The validation dataset primarily serves for Offline Policy Evaluation (OPE). For fairness, the experimental section of this paper exclusively uses the training datasets for all algorithms.

Specifically, the training datasets for Pipeline, Simglucose, RocketRecovery, and DMSD tasks each contain 100,000 samples. However, to simulate real-world data scarcity, the RocketRecovery training dataset has 19,977 samples, and SafetyHalfCheetah has 18,000. The Fusion task faces a particularly severe limitation, with its training dataset consisting of a mere 2,000 samples.The configuration of environments is summarized in Table \@\ref{task_configuration}. \tzl{A summary of the data descriptions based on these environmental features is detailed in the Table \@\ref{task_description}.}

\begin{table}[!h]
	\centering
        \renewcommand{\arraystretch}{1.3}
	\caption{Dataset description.}
	\label{task_description}
	\scalebox{0.9}{\begin{tabular}{c p{11cm}}
			\toprule
			Environment & Description \\
			\midrule           
                Pipeline & The delay set in the data is 25 steps. There are 20 trajectories in the dataset, each with a length of 1000 steps.  \\ 
                \midrule
                Simglucose & The delay in the data is influenced by many factors and is a dynamic value. The environment is affected by 30 external variables related to patient attributes.  \\ 
                \midrule
                RocketRecovery & The number of data trajectory is limited to only 100, with the longest trajectory having 500 steps and the shortest having 77 steps, averaging 200 steps. The data is influenced by external variables, specifically wind speed, which is not fixed; the distribution of wind speed in the data ranges from [-9.64, 9.64].  \\
                \midrule
                RandomFrictionHopper & The environment is influenced by external variables, particularly the friction coefficient, which is not fixed. The distribution of the friction coefficient in the data ranges from [1.50, 2.50], while the friction coefficient remains constant within a single trajectory.\\ 
                \midrule
                DMSD & The data was sampled using a PID control strategy. There are 200 trajectories in the dataset, each with a length of 100 steps.  \\ 
                \midrule
                Fusion &  The dataset only has limited 20 trajectories, with each entry having 100 steps.\\ 
                \midrule
                SafetyHalfCheetah & The global constraint in the environment is that the speed does not exceed 3.2096, and the data shows that the safe trajectory speed has not exceeded this upper limit. There are 18 trajectories in the dataset, each with a length of 1000 steps.  \\ 
			\bottomrule
	\end{tabular}}
\end{table}
\section{Computation Resources}\label{Appendix.Resources}

We utilized 16 NVIDIA RTX 4090 GPUs along with hundreds of CPU cores to conduct all experiments, employing Ray Tune \cite{2018raytune} for parallel training. Gathering the dataset took 6 hours, and training all the offline policies required approximately 2 weeks.
\section{Choice of Hyperparameters}\label{Appendix.HyperParams}

\begin{table}[!h]
	\centering
	\caption{The hyperparameter search space of model free algorithm.}
	\label{model-free-hyper-parameters_search_space}
	\scalebox{0.99}{\begin{tabular}{cc}
		\toprule
		Algorithms & Search Space\\
		\midrule
        BC & \begin{tabular}[c]{@{}c@{}}learning rate $\in \{1e-4, 5e-4, 1e-3\}$ \\ num\_layers $\in \{2, 3\}$\end{tabular} \\
		\midrule
		CQL & \begin{tabular}[c]{@{}c@{}}$\alpha \in \{5, 10\}$\\$\tau \in \{-1, 2, 5, 10\}$\\approximate-max backup $\in$ \{True, False\}\end{tabular}\\
		\midrule
        EDAC & \begin{tabular}[c]{@{}c@{}}num\_critics $\in \{10, 50\}$ \\$\eta \in \{1, 5\}$\end{tabular} \\
		\midrule
		MCQ & \begin{tabular}[c]{@{}c@{}}$\lambda \in \{0.3,0.4,0.5, 0.6, 0.7, 0.8, 0.9, 0.95\}$\\auto\_alpha $\in$ \{True, False\}\end{tabular}\\
		\midrule
		TD3BC & \begin{tabular}[c]{@{}c@{}}$\alpha \in \{0.05, 0.1, 0.2\}$\\policy noise $\in$ \{0.5, 1.5, 2.5\}\end{tabular}\\
		\bottomrule
	\end{tabular}}
\end{table}

For BC, we used the Adam optimizer during training, running 100K steps with a batch size of 256. To reduce overfitting, we implemented early stopping based on the minimum mean square error (MSE) implemented on the test dataset. Although the BC algorithm is relatively robust to parameter changes, we conducted parameter search training to ensure fairness. We particularly focus on the learning rate of the optimizer and the number of network layers, whose search ranges are $\{1e-4, 5e-4, 1e-3\}$ and $\{2, 3\}$, respectively. The experimental results further support the view that hyperparameters have a relatively small impact on the performance of the BC algorithm.

For CQL, we mainly consider 4 parameters mentioned in the original paper:
\begin{itemize}
	\item Q-values penalty parameter $\alpha$: In the formulation of CQL, $\alpha$ stands for how much penalty will be enforced on the Q function. As suggested in the paper, we search for $\alpha \in \{5, 10\}$.
	\item $\tau$: Since $\alpha$ can be hard to tune, the authors also introduce an auto-tuning trick via dual gradient descent. The trick introduces a threshold $\tau > 0$. When the difference between Q-values are greater than $\tau$, $\alpha$ will be auto-tuned to a greater value to make the penalty more aggressive. As suggested by the paper, we search $\tau \in \{-1, 2, 5, 10\}$. $\tau = -1$ indicates the trick is disable.
	\item Approximate-max backup: By default, the bellman backup is computed with double-Q, i.e. $y = r + \min_{i=1,2}Q_i(s^\prime, a^\prime)$, where $a^\prime \sim \pi(s^\prime)$. In addition, the authors propose approximate-max backup, which uses 10 samples to approximate the max Q-values, where the backup is computed by $y = r + \min_{i=1,2}\max_{a^\prime_1 ... a^\prime_{10} \sim \pi(s^\prime)}Q_i(s^\prime, a^\prime)$.
\end{itemize}

For EDAC, despite the original paper showing that the algorithm is not highly sensitive to hyperparameters, we still pay attention to the two parameters mentioned in the paper:
\begin{itemize}
	\item \texttt{num\_critics}: The number of critic networks in the ensemble plays a crucial role in the EDAC algorithm. Multiple critic networks are utilized to quantify the uncertainty in the critic value estimates. Increasing the number of \texttt{N} can enhance the algorithm's robustness against OOD (Out-Of-Distribution) data points, but it also raises the computational complexity. Based on the suggestions in the reference paper, we search num\_critics $\in \{10, 50\}$.
	\item $\eta$: This is the weight of the regularization term used to adjust the diversity of gradients within the set of critic networks. By minimizing the similarity between the gradients of different critic networks, the policy is encouraged not to be overly confident in OOD actions. The magnitude of $\eta$ directly affects the contribution of the gradient diversity regularization term to the overall loss function. Based on the suggestions in the reference paper, we search $\eta \in \{1, 5\}$.
 
\end{itemize}

For MCQ, we mainly consider 2 parameters mentioned in the original paper:
\begin{itemize}
         \item $\lambda$: This is a crucial hyperparameter used to balance the training of in-distribution actions and out-of-distribution (OOD) actions. The value of $\lambda$ directly affects the algorithm's emphasis on the standard Bellman error and OOD actions during training. If $\lambda$ is too small, it may lead to excessive focus on OOD actions, thereby impacting performance. As suggested by the paper, we search $\lambda \in \{0.3,0.4,0.5, 0.6, 0.7, 0.8, 0.9, 0.95\}$.
        \item \texttt{auto\_alpha}: This is a hyperparameter used for automatic entropy adjustment in the Soft Actor-Critic (SAC) algorithm. It controls the randomness and exploratory nature of the policy. As suggested in the paper, we search for \texttt{auto\_alpha} $\in$ \{True, False\}.
\end{itemize}

For TD3BC, we mainly consider 2 parameters mentioned in the original paper:
\begin{itemize}
	\item $\alpha$:  This is a hyperparameter used to determine the weighting of $\lambda$. $\lambda$ is used to balance the weight between value function maximization and behavior cloning. The value of $\alpha$ determines the scale of $\lambda$, thus influencing the relative importance of imitation learning (behavior cloning) and reinforcement learning (value function maximization) in policy updates. As suggested in the paper, we search for $\alpha \in \{0.05, 0.1, 0.2\}$.
	\item Policy noise: A parameter used in TD3 for exploration, it augments the diversity of the policy by adding noise to the policy outputs. The purpose of policy noise is to increase exploration to discover new and potentially better actions. As suggested in the paper, we search for policy noise $\in$ \{0.5, 1.5, 2.5\}.
\end{itemize}

\begin{table}[!h]
	\centering
	\caption{Hyperparameter search space for model training algorithms.}
	\label{model-hyper-parameters_search_space}
	\scalebox{0.99}{\begin{tabular}{cc}
		\toprule
		Algorithms & Search Space\\
		\midrule
		BC MODEL & \begin{tabular}[c]{@{}c@{}}transition scaler $\in$ \{True, False\}\\learning rate $\in \{1e-3, 3e-4\}$\\logvar-loss-coef $\in \{1e-2, 1e-3\}$\end{tabular}\\
		\bottomrule
	\end{tabular}}
\end{table}

To ensure fairness, we train all model-based algorithms using the same model. We employ a pre-training method called Behavior Cloning (BC) to train a model that is utilized by these algorithms. We also perform parameter search for the model to optimize its performance. We consider 3 parameters mentioned in the original paper:
\begin{itemize}
	\item transition scaler: This parameter is used to control whether normalization should be applied to the input. 
	\item learning rate: This parameter is used to set the learning rate of the optimizer. We search for learning rate $\in \{1e-3, 3e-4\}$.
	\item logvar-loss-coef: The role of this hyperparameter in the loss function is to balance and regulate the uncertainty range predicted by the model. Specifically, it ensures that the model maintains a reasonable prediction variance during the training process by controlling the contributions of the maximum and minimum logarithmic variances to the total loss. We search for logvar-loss-coef $\in \{1e-2, 1e-3\}$.
\end{itemize}

\begin{table}[!h]
	\centering
	\caption{The hyperparameter search space of model base algorithm.}
	\label{mdoel-base-hyper-parameters_search_space}
	\scalebox{0.99}{\begin{tabular}{cc}
		\toprule
		Algorithms & Search Space\\
		\midrule
		MOPO & \begin{tabular}[c]{@{}c@{}}uncertainty type $\in$ \{aleatoric, disagreement\}\\$h \in \{1, 5\}$\\$\lambda \in \{0.5, 1, 2, 5\}$\end{tabular}\\
		\midrule
		COMBO & \begin{tabular}[c]{@{}c@{}}conservative loss data type $\in$ \{model, mix\}\\$h \in \{1, 5\}$\\cql loss weight $\in \{0.5, 1, 2, 5\}$\end{tabular}\\
		\midrule
		RAMBO & \begin{tabular}[c]{@{}c@{}}$\lambda \in \{0, 3e-4, 1e-3\}$\\$h \in \{1, 5\}$\\adv\_lr $\in$ \{1e-3, 3e-4\}\end{tabular}\\
		\midrule
		MOBILE & \begin{tabular}[c]{@{}c@{}}$h \in \{1, 5\}$\\penalty coefficient $\in \{0.5, 1.5, 2.5, 3.5\}$\end{tabular}\\
		\bottomrule
	\end{tabular}}
\end{table}

For MOPO, we consider 3 parameters mentioned in the original paper:
\begin{itemize}
	\item Uncertainty type: In the default setting, MOPO uses the max L2-norm of the output standard deviation among ensemble transition models, i.e. $\max_{i=1...N}\|\sigma_\theta^i(s, a)\|_2^2$, as the uncertainty term. Since the learned variance can theoretically recover the true aleatoric uncertainty, we denote this type of uncertainty as aleatoric. We also include another variant that uses the disagreement between ensemble transition models, i.e. $\max_{i=1...N}\|\Delta_\theta^i(s, a) - \frac{1}{N}\sum_i \Delta_\theta^i(s, a)\|_2^2$, as the uncertainty term. We refer to this variant as disagreement.
	\item $h$: MOPO uses a branch rollout trick that run rollout from states in the dataset with a small length. $h$ determines the length of the rollout. As suggested in the paper, we search for $h \in \{1, 5\}$.
	\item $\lambda$: The main idea of MOPO is to penalize the reward function with the uncertainty term, i.e. $\hat{r} = r - \lambda u(s, a)$. Here, $\lambda$ controls the amplitude of the penalty. As suggested in the original paper, we search for $\lambda \in \{0.5, 1, 2, 5\}$.
\end{itemize}

For COMBO, we consider 3 parameters mentioned in the original paper:
\begin{itemize}
	\item Conservative loss data type: COMBO introduces a conservative regularization loss term to suppress overly optimistic value estimates. This parameter controls whether the data used to compute this loss term is generated solely by the model (model) or a mixture of model-generated data and offline expert data (mix).
	\item $h$: Like MOPO, COMBO uses a branch rollout trick that run rollout from states in the dataset with a small length. $h$ determines the length of the rollout. As suggested in the paper, we search for $h \in \{1, 5\}$.
	\item Cql loss weight: COMBO combines MOPO and CQL algorithms. This hyperparameter is used to adjust the contribution of conservative loss to total loss in the CQL algorithm. As suggested in the original paper, we search for $\lambda \in \{0.5, 1, 2, 5\}$.
\end{itemize}

For RAMBO, we consider 3 parameters mentioned in the original paper:
\begin{itemize}
	\item $\lambda$:This is a hyperparameter used to balance the trade-off between model prediction accuracy and policy conservatism. $\lambda$ is used to adjust the trade-off between maximizing the maximum likelihood estimation loss on the distribution within the dataset and minimizing the loss on reducing the policy value function across the entire model parameter space. As suggested in the paper, we search for $\lambda \in \{0, 3e-4, 1e-3\}$.
	\item $h$: This parameter determines the length of the k-step rollout used by the model to generate synthetic data. It determines the number of future steps considered when simulating future trajectories, which impacts the foresight in policy evaluation and optimization. As suggested in the paper, we search for $h \in \{1, 5\}$.
	\item adv\_lr : RAMBO introduces an adversarial environment model to enhance the robustness of the policy. Specifically, RAMBO formulates the offline reinforcement learning problem as a two-player zero-sum game, where one player is the policy-optimizing agent and the other player is the adversarial environment model. This parameter refers to the learning rate of the optimizer used for adversarial training. As suggested in the paper, we search for $\lambda \in \{1e-3, 3e-4\}$.
\end{itemize}

For MOBILE, we consider 3 parameters mentioned in the original paper:
\begin{itemize}
	\item $h$: This parameter represents the number of steps used for model-based rollouts to generate synthetic data. A shorter rollout length may reduce prediction uncertainty, while a longer rollout can provide more information about future states but may also increase uncertainty. As suggested in the paper, we search for $h \in \{1, 5\}$.
	\item Penalty coefficient: This is the coefficient used to adjust the penalty term in the Bellman estimate. In the MOBILE algorithm, the penalty coefficient is used to balance the reliability of model predictions and the conservatism of the policy. A larger value of the penalty coefficient increases the penalty on model uncertainty, resulting in a more conservative policy. As suggested in the original paper, we search for $\lambda \in \{0.5, 1.5, 2.5, 3.5\}$.
\end{itemize}

\begin{figure}
    \centering
    \includegraphics[width=1\linewidth]{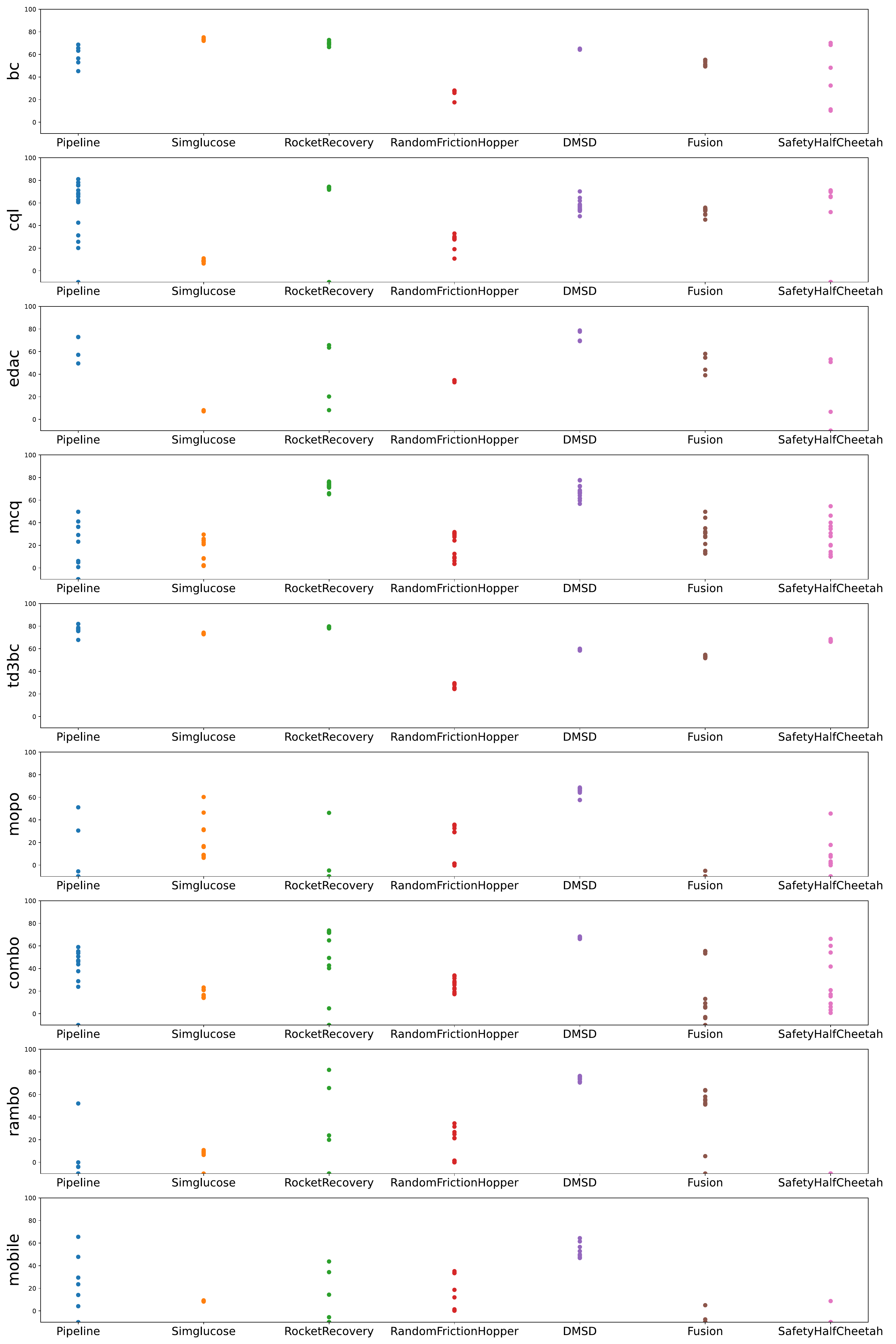}
    \caption{The results of all hyperparameter combinations. To enhance the display effect, all scores below -10 are clipped to -10.}
    \label{fig:hyperparameter_sensitivity}
\end{figure}

Figure \@\ref{fig:hyperparameter_sensitivity} shows the results of all hyperparameter combinations. Table \@\ref{hyper-parameters_reported_results} summarizes the hyperparameters used in the reported results of the paper.

\begin{table}[htbp]
\centering
\caption{Hyper-parameters for reported results.}
\label{hyper-parameters_reported_results}
\renewcommand{\arraystretch}{1.5}
\resizebox{\linewidth}{!}{%
\begin{tabular}{lllllllll}
\toprule
 Algorithm & Hyper-parameters & Pipeline & Fusion & DMSD & RocketRecovery & Simglucose & SafetyHalfCheetah & RandomFrictionHopper \\
\midrule
\multirow[t]{2}{*}{BC} & learning rate & 1e-3 & 5e-4 & 1e-3 & 1e-4 & 1e-3 & 1e-3 & 1e-3 \\
 & num\_layers & 3 & 3 & 2 & 2 & 3 & 3 & 2 \\
\cline{1-9}
\multirow[t]{3}{*}{CQL} & $\alpha$ & 10 & 10 & 5 & 10 & 10 & 10 & 10 \\
 & $\tau$ & 10 & 5 & 10 & 2 & 2 & 2 & 5 \\
 & approximate-max backup & 2 & 2 & 3 & 3 & 2 & 2 & 2 \\
\cline{1-9}
\multirow[t]{2}{*}{EDAC} & num\_critics & 10 & 50 & 10 & 50 & 10 & 50 & 10 \\
 & $\eta$ & 5 & 5 & 1 & 1 & 1 & 1 & 1 \\
\cline{1-9}
\multirow[t]{2}{*}{MCQ} & $\lambda$ & 0.9 & 0.95 & 0.95 & 0.4 & 0.5 & 0.3 & 0.9 \\
 & auto\_alpha & False & True & False & False & True & True & True \\
\cline{1-9}
\multirow[t]{2}{*}{TD3BC} & $\alpha$ & 0.2 & 0.2 & 0.2 & 0.2 & 0.2 & 0.05 & 0.2 \\
 & policy noise & 0.5 & 1.5 & 0.5 & 1.5 & 2.5 & 2.5 & 0.5 \\
\cline{1-9}
\multirow[t]{2}{*}{MOPO} & uncertainty type & aleatoric & aleatoric & aleatoric & aleatoric & aleatoric & aleatoric & aleatoric \\
 & $h$ & 5 & 5 & 1 & 5 & 1 & 5 & 5 \\
 & $\lambda$ & 5 & 2 & 0.5 & 5 & 5 & 5 & 0.5 \\
\cline{1-9}
\multirow[t]{3}{*}{COMBO} & cql loss weight & 5 & 5 & 2.5 & 5 & 2.5 & 5 & 3.5 \\
 & $h$  & 5 & 5 & 1 & 1 & 5 & 1 & 1 \\
 & conservative loss data type & model & model & mix & model & model & model & mix \\
\cline{1-9}
\multirow[t]{2}{*}{RAMBO} & $\lambda$ & 3e-4 & 3e-4 & 1e-3 & 0 & 0 & 3e-4 & 0 \\
 & $h$ & 5 & 1 & 1 & 5 & 5 & 1 & 5 \\
 & adv\_lr & 1e-3 & 1e-3 & 1e-3 & 1e-3 & 1e-3 & 3e-4 & 1e-3 \\
\cline{1-9}
\multirow[t]{3}{*}{MOBILE} & $h$ & 1 & 5 & 1 & 5 & 1 & 5 & 5 \\
 & penalty coefficient & 3.5 & 1.5 & 0.5 & 3.5 & 2.5 & 0.5 & 1.5 \\
\cline{1-9}

\end{tabular}%
}
\end{table}

\section{Comparison of Response Curves}

\begin{figure}
\centering 
\includegraphics[width=1\textwidth]{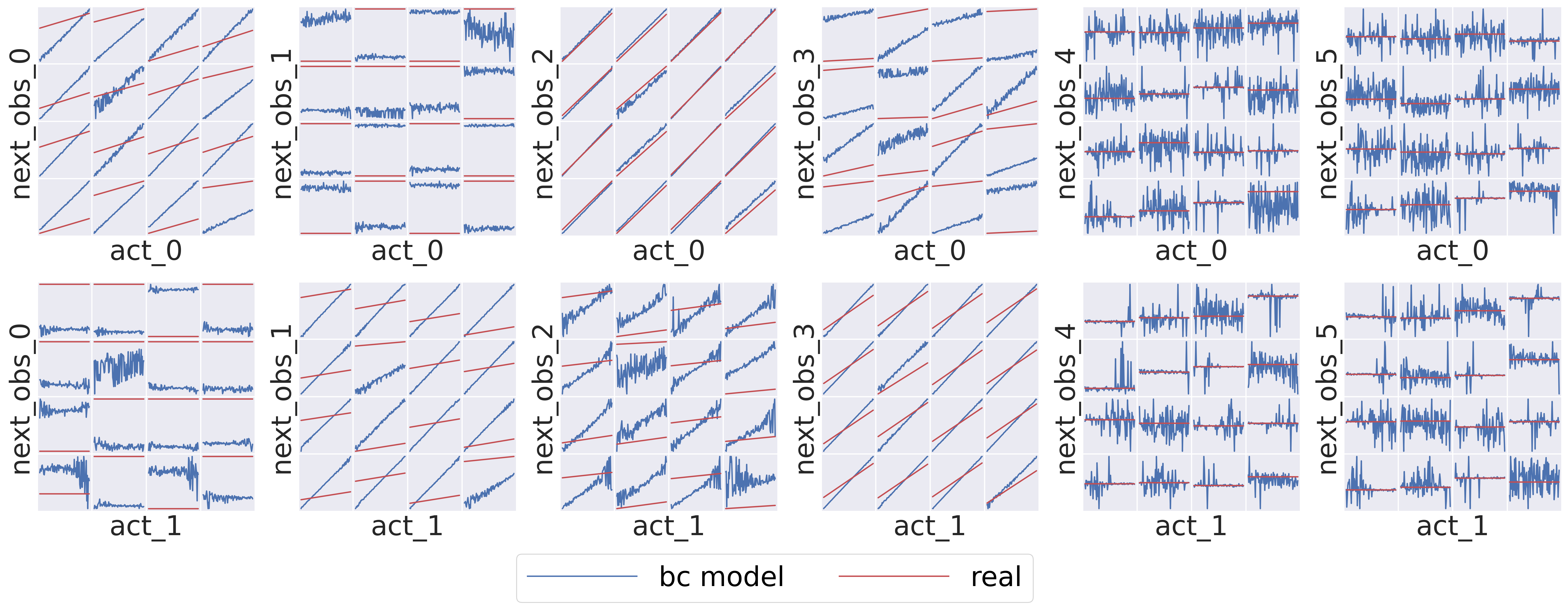} 
\caption{Comparison of response curves of dynamic model and real environments for DMSD.}
\label{fig.responsecurve} 
\end{figure}

Generally, model-based algorithms use BC to learn the dynamic model. Figure~\ref{fig.responsecurve} compares the dynamic model in the DMSD with the real environment. We did this by first randomly selecting 16 sets of observation data from the dataset, resulting in each subplot containing 16 plots. We then sampled the action space within the range $[-1,1]$ for two dimensions of actions. These initial observations and actions were fed into the transition model to obtain the next state. The transitions predicted by the model are marked with blue lines in the figure. The transitions in the real DMSD environment are marked with red lines.

From the figure, we can observe that while the dynamic model can approximate the real model under the influence of some actions in certain dimensions, in most dimensions, the dynamic model only aligns with the trend of the line of the real model. It does not accurately match the real model. Although the baseline algorithms' final results exceed the data scores in the DMSD task, the transition models in the model-based algorithms do not provide accurate transitions through the BC method. This might be due to the offline dataset collected through the traditional control method (e.g. PID), which tends to make the transition model learned by BC hard to identify the correct causal relationship of the states and actions~\citep{2023chen}.

\end{document}